\documentclass[conference]{IEEEtran}
\IEEEoverridecommandlockouts

\usepackage[utf8]{inputenc} % allow utf-8 input
\usepackage[T1]{fontenc}    % use 8-bit T1 fonts
\usepackage{hyperref}       % hyperlinks
\usepackage{url}            % simple URL typesetting
\usepackage{booktabs}       % professional-quality tables
\usepackage{amsfonts}       % blackboard math symbols
\usepackage{nicefrac}       % compact symbols for 1/2, etc.
\usepackage{microtype}      % microtypography
\def\BibTeX{{\rm B\kern-.05em{\sc i\kern-.025em b}\kern-.08em
    T\kern-.1667em\lower.7ex\hbox{E}\kern-.125emX}}
\usepackage{graphicx}
\usepackage{subcaption}
\usepackage{csvsimple}
\usepackage{rotating}
\usepackage{cite}
\usepackage{tabularx}
\usepackage[table]{xcolor}
\newcommand{\citep}{\cite} % to be compatible with bibliographystyle like citation

\usepackage{snaptodo}
\usepackage{makecell}
\usepackage{enumitem}

\snaptodoset{block rise=1em}
\snaptodoset{margin block/.style={font=\tiny}} 
 \snaptodoset{margin block/.style={font=\tiny}} 

\snaptodoset{margin block/.style={font=\tiny}}

\usepackage{cite}
\usepackage{amsmath,amssymb,amsfonts}
\usepackage{algorithmic}
\usepackage{graphicx}
\usepackage{textcomp}
\usepackage{xcolor}
\usepackage{subcaption}

\def\BibTeX{{\rm B\kern-.05em{\sc i\kern-.025em b}\kern-.08em
    T\kern-.1667em\lower.7ex\hbox{E}\kern-.125emX}}
\begin{document}

\title{Vectorizing string entries for data processing on tables: when are
larger language models better?
}

\author{\IEEEauthorblockN{1\textsuperscript{st} Léo Grinsztajn}
\IEEEauthorblockA{\textit{SODA} \\
\textit{INRIA}\\
leo.grinsztajn@inria.fr}
\and
\IEEEauthorblockN{2\textsuperscript{nd} Myung Jun Kim}
\IEEEauthorblockA{\textit{SODA} \\
\textit{INRIA}}
\and
\IEEEauthorblockN{3\textsuperscript{rd} Edouard Oyallon}
\IEEEauthorblockA{\textit{MLIA} \\
\textit{CRNS, Sorbonne University}}
\and
\IEEEauthorblockN{4\textsuperscript{th} Gaël Varoquaux}
\IEEEauthorblockA{\textit{SODA} \\
\textit{INRIA}}
}

\maketitle

\begin{abstract}
There are increasingly efficient data processing pipelines that work on
vectors of numbers, for instance most machine learning models, or vector
databases for fast similarity search. These require converting the data
to numbers. While this conversion is easy for simple numerical and
categorical entries, databases are strife with text entries, such as names or descriptions.
In the age of large language models, what's the best strategies to vectorize tables entries, baring in mind that larger models entail more operational complexity?
We study the benefits of language models in 14 analytical tasks on tables while varying the training size, as well as for a fuzzy join benchmark. 
We introduce a simple characterization of a column that reveals two settings:
\emph{1)} a \emph{dirty categories} setting, where strings share much similarities
across entries, and conversely \emph{2)} a \emph{diverse entries}
setting. For dirty categories, pretrained language models bring
little-to-no benefit compared to simpler string models. For diverse
entries, we show that larger language models improve data processing. For
these we investigate the complexity-performance tradeoffs 
and show that they reflect those of classic
text embedding: larger models tend to perform better, but it is useful
to fine tune them for embedding purposes.
\end{abstract}


\section{Introduction}

While much of data engineering deals with discrete entries --categories,
normalized entities, or open-ended text-- there is a growing trend to use
data representations made of numerical vectors. For instance, vector
databases \cite{hanComprehensiveSurveyVector2023} use such representations in fast similarity searches for
retrieval and fuzzy joins. Neural networks, which brought revolutions in
many aspects of data processing, are also based on numerical vectors to
represent the available information, including in natural language
applications which deal solely with discrete tokens. However, for typical
data tables, with columns containing entries of different nature and type, recent work has shown that bigger, more sophisticated, neural methods do not outperform simpler machine-learning models based on trees \cite{grinsztajnWhyTreebasedModels2022}. These tree-based methods handle discrete entries naturally, but struggle when the data cannot be represented as a moderate number of categories. In such a case, it is useful to combine them with representations of the string surface form of the entries \cite{cerdaEncodingHighcardinalityString2022}.

% text embeddings in general, and say that in our case it's not obvious we need them because we have short strings
Good vectorial representation of the string entries in tables remains
crucial. Practitioners often rely on pretrained word embeddings
developed in natural language processing \cite{joulinBagTricksEfficient2017} or
numerical representations built from substrings
\cite{cerdaEncodingHighcardinalityString2022}. Modern natural language
processing has moved on to much more elaborate architectures, using
pretrained attentional architectures \cite{devlinBERTPretrainingDeep2019} which have
evolved to large language models LLMs, such as LLaMa
\cite{touvronLLaMAOpenEfficient2023}. But vectorizing text with
very large language models requires multiple expensive and
rare high-end GPUs due to their memory
footprint; it induces large energy consumption 
\cite{luccioniPowerHungryProcessing2023}. By contrast, table entries are typical fairly short strings.
They seldom have the complex grammatical or narrative structures that
pushed the development of language models of increasing depth and context
window. This beg the question: what are the computational trade-off to create
vectorial representations of string entries in tables? Are pretrained
language models needed or are string representations enough? How complex should a model be? Given the cottage industry of language model
--to date, the HuggingFace model hub has 42\,000 models for text
classification, 2\,700 for sentence embedding--, which one to choose to
embed text entries in tables? Evaluating many models for a given
analysis is clearly impracticable; there is a dire need for guidelines.

Here we contribute a thorough empirical study of embedding of string
entries in table for data processing. We consider two settings: \emph{1)}
Data analytics, \emph{ie} statistical analysis of records in a table, where we
consider 14 supervised learning tasks, and \emph{2)} Data engineering, in
particular table assembly, where we consider fuzzy-join: joining across
50 pairs of tables with imperfect alignment in the entity surface forms.
We investigate more than 30 string embedding approaches. We show that a
simple measure of the diversity across string entries enables separating
columns on which string representations suffice, with entries that
resemble ``\emph{dirty categories}'', and columns with more \emph{diverse
entries} on which large language models are beneficial. On the diverse
entries, we show that the learnings from the text-embedding literature in
natural language processing carry over to the data engineering settings.

Section \ref{sec:context} introduces the specific problem settings that we
study and the related work on embedding entries. Section \ref{xp} then
describes our benchmarking material: the datasets we use and the
embedding methods that we survey. Finally, \autoref{results} details the
results from the benchmark, highlighting various important trends, before
we conclude in \autoref{sec:conlcusion}, giving high-level
recommendations to encode text entries for data processing.

\bigskip
\section{Context and related works}
\label{sec:context}

\subsection{Problem setting: vectorization in data processing}
\label{problem_setting}

\paragraph*{Analytics}
Analytical tasks on tables tackle, in general, estimation of statistical
properties of the records (entries in a row). Often these properties
are conditional estimates of
one attribute in a row as a function of others; 
For instance, in a real-estate application, one might be interested in
linking the expected price of properties to their features, such as age,
number of rooms... Such estimations can be
cast in a statistical learning framework \cite{cvetkov-ilievRelationalDataEmbeddings2022}.
The statistical estimation is formulated on a dataset of $n$ observations
$(x_1, y_1), (x_2, y_2), ..., (x_n, y_n)$, where each observation
consists of a feature vector $x_i \in \mathbb{R}^p$, the input
attributes, and an outcome $y_i
\in \mathbb{R}$ or $y_i \in {1, 2, ..., K}$, the target attribute. For practitioners, however,
this setting typically only appear toward the end of a long data
engineering process. First, text and categorical features must be
vectorized, which is especially challenging for high-cardinality
categorical features. Second, information is often distributed across
multiple tables, and a time-consuming part of the data processing
pipeline consists of carefully joining these different tables. This paper
focuses on the text entries, which lead to significant challenges in the
data processing operation.
It explores a pipeline based on vectorizing these text entries prior to
statistical learning or joining tables. A good embedding approach is one that makes
downstream tasks --predictions, joins-- more accurate.

\paragraph*{Fuzzy join}
Fuzzy join, --and the related similarity-join, fuzzy-matching, and entity
resolution--, requires linking across different tables entries which refer to the same entity. We focus on the many-to-one join problem, where we want to enrich a base table with an auxiliary table (the reference table). More formally, as described in \cite{liAutoFuzzyJoinAutoProgramFuzzy2021}, if we denote $L$ and $R$ two input tables, where $L$ serves as the reference table, a fuzzy join can be  defined as a function $J : R \rightarrow L \cup \bot$, where $\bot$ denote no match. Note that each element from $R$ can match only one element in $L$, the reference table, while each elements of $L$ can match many elements in $R$.

Fuzzy joining often makes use of Nearest Neighbor algorithms on a well chosen representation of the data. As for data analytics, we study a simple pipeline, were we vectorize text entries prior to using a Nearest Neighbor algorithm. A good embedding should make the downstream matching more accurate.

\paragraph*{Vectorizing records}
For both tasks, analytics with statistical learning and fuzzy joining, we investigate a
simple tabular data processing pipeline: text and high-cardinality
features are vectorized using a language model (and concatenated to the numerical features for tabular analytics) and fed into a classical machine learning model. While each ad-hoc module results from a complex learning process, their aggregation into a tabular data processing pipeline is straightforward.

Vectorizing can be applied offline, prior to data analysis, 
as it is computed row by row, and the
resulting feature engineering can be reused across many analytical task.
Such a reuse simplifies operations and decreases computational costs. But
it must be put in perspective with the operational costs of the embedding
model.

\smallskip
\subsection{Related work: many ways to represent table entries}

\paragraph*{Encoding high-cardinality features}
Given a table with text entries, the traditional statistical literature
often relies on One-Hot Encoding, but it falls short when dealing with
high-cardinality categories, as in creates an explosion of the
dimensionality of resulting embeddings. To alleviate the problem,
various replacement methods have been suggested. Target Encoding is a
competitive alternative that associates each category with the average
value of the target variable \cite{micci-barrecaPreprocessingSchemeHighCardinality2001}, but it breaks when dealing with categories
not seen during the training (out-of-vocabulary problem). 

Character-level approaches based on substrings can generalize to unseen text and improve data processing tasks
\cite{cerdaEncodingHighcardinalityString2022}.
%\gv{TODO (likely for Gael); Here explain quickly the idea behind count-vectorizer and minhashencoding} 
A central idea here is to count occurrences of sub strings, for instance
defined by words or character-level n-grams. These counts can then be turned
into low-dimensional embeddings with a matrix factorization, for instance
a PCA after Tf-Idf renormalization (term frequency–inverse document
frequency) to make the count distributions more suited for the square
loss. A more advanced approach, yet fast and lightweight relies on
MinHash sketching --a probabilistic approach to capturing Jaccard
similarities between sub-string ensembles-- to create embeddings that
expose containment \cite{cerdaEncodingHighcardinalityString2022}.
Sub-string level models are widely used as part of machine-learning software packages such as Scikit-Learn
\citep{scikit-learn} or Skrub \cite{skrub2023}. These approaches, however, can only rely on the regularity in the data, as they do not incorporate any outside semantic information.

\paragraph*{Incorporating external information}
Enhancing tabular data with external information, often referred to as feature enrichment, can significantly boost the prediction accuracy. If done manually, however, this process typically requires intensive labor from skilled data scientists, often involving painful joins and aggregations. To automate the process, Deep Feature Synthesis \cite{kanterDeepFeatureSynthesis2015} greedily carries out joins and aggregations across tables. However, it is not applicable on large databases where it faces tractability challenges and results in extremely high-dimensional vectors.

To mitigate this issue, subsequent research has attempted to generate useful embeddings for entities within tabular data. \cite{cvetkov-ilievRelationalDataEmbeddings2022} developed a method that learns embeddings from knowledge graphs. They demonstrated that such embeddings brings background information that enhances performance when incorporated into various tables. However, this approach requires a challenging step involving explicitly matching text entries between tables and knowledge graphs.

\paragraph*{Language models for tabular data prediction}

With the widespread use of language models, several works have been proposed to enhance predictions for tabular data. Given that they are trained on huge corpora of texts, the embeddings from the language models can provide useful background knowledge. For example, \cite{carballoTabTextFlexibleContextual2023} observed that performance improved on one clinical dataset when using BERT-embeddings. Similarly, \cite{cerdaEncodingHighcardinalityString2022} reported competitive results when employing this approach. Moreover, language models are robust to variations in text entries \cite{chenImputing2022}, which solves the issue of rigorous entity matching required when incorporating external information. 

Additionally, several works extend the use of language models beyond embedding entities to enhance predictions. \cite{hollmannLargeLanguageModels2023a} leverages recent advancements in code generation with language models to automatically generate new features, retaining only those that boost performance. \cite{hegselmannTabLLMFewshotClassification2023} and \cite{dinhLIFTLanguageInterfacedFineTuning2022} directly fine-tune a language model on raw data, reporting good performance on very small datasets. These models rely both on the background knowledge and predictive abilities of language models, making it challenging to disentangle their respective contributions. In this work, we show how language models can bring in background information, as opposed to string models learned on the table at hand.

\paragraph*{Probing}
Starting with \cite{alainUnderstandingIntermediateLayers2018}, researchers have been training simple models on intermediary activation of neural networks to uncover the information contained in these hidden states. The motivation of this line of work is often to better understand the inner workings of these models. In this paper, we use similar methods for a more practical aim: to easily extract vectorized information from textual entities. More closely related to our work, \cite{gurneeLanguageModelsRepresent2023} shows that probing methods can extract detailed information about the spatial and temporal location of entities from large language models such as LLaMA2 \citep{touvronLlamaOpenFoundation2023}.

\paragraph*{Text embeddings}

Sentence embeddings provide a compact way to represent a text and its information. For this reason, they are now used for various purposes, from text classification to paragraph retrieval.

While such embeddings can be directly extracted from language models pretrained on pretext tasks, \cite{liSentenceEmbeddingsPretrained2020} argues that the semantic information inside the model embeddings is not fully exploited without finetuning.
%TODO: other paper claiming that cosine_sim doens't work well without finetuning
This has lead to a rich line of research on finetuning methods for sentence embeddings, using various methods such as constrastive training \cite{gaoSimCSESimpleContrastive2022}\cite{niSentenceT5ScalableSentence2021}\cite{neelakantanTextCodeEmbeddings2022}\cite{wangTextEmbeddingsWeaklySupervised2022}, finetuining for classification on labeled sentence pairs datasets such as NLI or NQ \cite{kwiatkowskiNaturalQuestionsBenchmark2019}\cite{wangTextEmbeddingsWeaklySupervised2022} or training to imitate slower but better-performing cross encoder models \cite{thakurAugmentedSBERTData2021}, which take a pair of sentence as input.

While  these models and methods have been evaluated on various tasks \cite{muennighoffMTEBMassiveText2023}, they have not been studied in the specific context of tabular data processing and analytics, where string entries are typically quite short and redundant, free-form text is scarce, and text embeddings are sometimes combined with numerical features.

%TODO LLMs: explain the difference between the ones you probe

\paragraph*{Table models}
Following pretrained-language models, the training scheme of these models have been tailored to inputs belonging to tables, leading to pretrained table models \cite{dengTURLTableUnderstanding2020} \cite{zhangTableLlamaOpenLarge2023}. Compared to their text-trained counterparts, these models have shown improved performances on table specific tasks such as row population, entity linking, or table fact verification. In this paper, we do not directly use models to solve table specific tasks, but rather attempt to vectorize table entries to improve performance on data analytics and preprocessing.

\bigskip
\section{Experimental setup: probing analytics and joins}
\label{xp}

\subsection{An analytics benchmark: predicting an attribute value}

To evaluate the performance of different text entries vectorization schemes for tabular analytics, we start by introducing a new classification benchmark on datasets containing both useful numerical features and text entries.

\paragraph{Datasets}
We gathered datasets across multiple sources, mainly previous machine learning studies and kaggle competitions. Most machine-learning studies unfortunately focus on numerical data and we found 28 tabular 
datasets with at least one of the column being a text entry and with at least 1500 rows. Out these, 13 datasets (14 tasks) have at least one string column that is important for prediction \footnote{On the 28 datasets we consider, 11 show ROC-AUC gains of less than 1\% when including the text features, compared to using only the numerical features, and 14 show gains of less that 3\%. These gains are computed by taking the biggest gains among OpenAI embeddings, Skrub MinHashEncoder, and the 3 best models in the MTEB benchmark.  We restrict our analysis to the 14 datasets with gains greater than 3\%.}. The text features contained in these tables are diverse, as shown in table \ref{table:examples}.:
\begin{enumerate}
    \item \textbf{Bikewale} \cite{magellandata} \footnote{http://pages.cs.wisc.edu/\textasciitilde anhai/data/784\_data/bikes/csv\_files/bikewale.csv} Information on bikes and scooters in India. The task is to predict the degree of price of automobiles.
    \item \textbf{Clear Corpus} \cite{crossleyLarge2023}\footnote{https://www.commonlit.org/blog/introducing-the-clear-corpus-an-open-dataset-to-advance-research-28ff8cfea84a/}: Generic information about the reading passage excerpts for elementary school students. The task is to predict the readability of the excerpts. The text feature is the name of the book, not the excerpt.
    \item \textbf{Company Employees}\footnote{https://www.kaggle.com/peopledatalabssf/free-7-million-company-dataset}: Information on companies with over $1,000$ employees. The task is to predict the size range of the companies.
    \item \textbf{Employee Salaries} \footnote{https://openml.org/d/42125}: Information on salaries for employees of the Montgomery County, MD. The task is to predict the current annual salary range of the employees.
    \item \textbf{Employee remuneration and expenses earning over 75000}  \footnote{\url{https://opendata.vancouver.ca/explore/dataset/employee-remuneration-and-expenses-earning-over-75000/information/?disjunctive.department&disjunctive.title}} Remuneration and expenses for employees earning over \$75,000 per year. The task is to predict the remuneration of employees.
    \item  \textbf{Goodreads} \cite{magellandata} \footnote{http://pages.cs.wisc.edu/\textasciitilde anhai/data/784\_data/books2/csv\_files/goodreads.csv} Datasets containing information about books. The task is to predict the average rating of each book.
    \item  \textbf{Journal Influence}: Scientific journals and their descriptive features. The task is to predict the influence of a journal. 
    \item \textbf{Spotify}\footnote{https://www.kaggle.com/datasets/maharshipandya/-spotify-tracks-dataset}: Generic information on Spotify tracks with some associated audio features. The task is to predict the popularity of the albums.
    \item \textbf{US Accidents}\footnote{https://smoosavi.org/datasets/us\_accidents}: Information of accidents in US cities between 2016 and 2020. From this dataset, two tasks are conducted: (1) the range of accident counts for the US cities (2) the severity of the reported accidents.
    \setcounter{enumi}{10}
    \item \textbf{US Presidential} \cite{cvetkov-ilievRelationalDataEmbeddings2022}: Voting statistics in the 2020 US presidential election along with information on US counties. The task is to predict the range of voting numbers across US counties.
    \item \textbf{Ramen ratings}\footnote{https://www.kaggle.com/datasets/residentmario/ramen-ratings}. The dataset contains ratings and characteristics of various ramens produced from multiple countries. The task is to predict the ratings of the ramens.
by the Chicago Department of Buildings since 2006. The task is to predict the Total Fee.
    \item \textbf{Wine reviews} \cite{deazambujaXWinesWineDataset2023} \footnote{https://github.com/rogerioxavier/X-Wines} The dataset contains wine ratings, as well as various such as price, winery or a small description. The task is to predict the rating.
    \item \textbf{Zomato}\footnote{https://www.kaggle.com/datasets/himanshupoddar/zomato-bangalore-restaurants}. Information and reviews of restaurants in Bengaluru, India. The task is to predict the ratings of the restaurants.    
\end{enumerate}

\begin{table}[]
{\small
\rowcolors{2}{gray!25}{white}
\begin{tabular}{p{1.3cm}p{1.2cm}p{3.7cm}p{1cm}}
\rowcolor{gray!50}
\textbf{Dataset} & \textbf{Column} & \textbf{Example} & \textbf{Ngrams / 1000 rows} \\
Wine Review & Country & Portugal & 294 \\
Bikewale & Bike Name & Honda CB Twister Drum/Electric start & 2878 \\
Zomato & Location & Koramangala 1st Block & 1121 \\
Zomato & Name & Tandoor Garden & 8491 \\
Zomato & Dish Liked & Kaju Katli, Gulab Jamun, Petha & 7595 \\
Employee Salary & Department Name & Fire and Rescue Services & 932 \\
Spotify & Song name & She's So Mellow & 9688 \\
Spotify & Artist name & Brandtson & 12605 \\
Company Employees & Domain name & bajajfinserv.in & 8936 \\
Company Employees & Industry & food \& beverages & 2003 \\
Journal Influence & Journal name & Acta Biomaterialia & 8088 \\
Goodreads & Description & Anarchist, journalist, drama critic, advocate of birth control and free love, Emma Goldman was the most famous-and notorious-woman in… & 66423 \\
Ramen Ratings & Brand & Sapporo Ichiban & 3120 \\
Ramen Ratings & Variety & Tom Yum Seafood Creamy & 8975 \\
\end{tabular}
}
\caption{Examples of text features in our datasets.}
\label{table:examples}
\end{table}

\paragraph{Text and numerical features processing}
We consider a feature to be a text feature if its cardinality is greater than thirty. Other features (low cardinality categorical, numerical features and datetime features) will be referred as "numerical features" for simplicity, and are vectorized independently. We use a OneHotEncoder for low cardinality variables, MinHashEncoder \cite{cerdaEncodingHighcardinalityString2022} for features with a cardinality greater than 10, and the DatetimeEncoder from the package Skrub \cite{skrub2023} for datetime features (it transforms the datetime into features corresponding to the year, month, day, hour etc.). Numerical features are scaled with scikit-learn's \cite{scikit-learn} StandardScaler. Regression datasets are converted to binary classification, and all dataset are balanced. Except specified otherwise, we use sklearn's GradientBoostingClassifier as a classifier, as it is a strong baseline \cite{grinsztajnWhyTreebasedModels2022}. The same model is used on the text embeddings combined with numerical features. For a discussion of this choice, see \ref{simple_pipeline}.

\paragraph{Evaluations}
We use the same sample size for all datasets. This size varies accross experiments, as specified, and we limit ourselves to sample sizes below 5000, which still encompass a large part of the datasets used by practitioners \cite{LargestDatasetAnalyzed}. Evaluations are always done on 7 cross-validation folds.

\bigskip
\subsection{Entity resolution: A fuzzy join benchmark}
\label{fuzzy_join}
We also investigate embeddings in the context of a common time-consuming data processing step: entity resolution. More precisely, we focus on the many-to-one fuzzy join problem, which is emblematic of situations where we aim to enrich a base table with auxiliary tables containing more detailed information, as described in \ref{problem_setting}.

\paragraph*{Datasets}
For benchmarking, we take the 50 pairs of tables from \cite{liAutoFuzzyJoinAutoProgramFuzzy2021}. These dataset pairs are constructed using multiple snapshot from Wikipedia, and using the natural variations in page names to get different names for similar entities.

\paragraph*{Method}
Our simple pipeline consists of using a 1-NearestNeighbor on vectorized representations of the rows. These representation are computed using language model embeddings, or using scikit-learn's TfidfVectorizer for comparison. We also compare this pipeline to AutoFuzzyJoin \cite{liAutoFuzzyJoinAutoProgramFuzzy2021}, a state-of-the-art unsupervised framework that can
infer suitable fuzzy-join programs on given input tables. Note that for benchmarking, we use the datasets introduced in the AutoFuzzyJoin paper \cite{liAutoFuzzyJoinAutoProgramFuzzy2021}.

\smallskip
\subsection{Text embedding methods surveyed}
\label{text_embeddings}
\paragraph{Language models}
We aim to evaluate diverse language models. We first gathered models from the top of the MTEB benchmark \footnote{As indicated by this leaderboard around November 2023: https://huggingface.co/spaces/mteb/leaderboard} \cite{muennighoffMTEBMassiveText2023}. In particular, we focus on the two models at the top for some experiments:
\begin{itemize}
    \item BAAI's bge-large-en-v1.5 \cite{xiaoCPackPackagedResources2023} \footnote{https://huggingface.co/BAAI/bge-large-en-v1.5}: a 335M parameters model pretrained on a large scale corpus, and finetuned on corpuses of text pairs.
    \item LLMrails's ember-v1 \footnote{https://huggingface.co/llmrails/ember-v1}: a 335M parameters model trained on an extensive corpus of text pairs.
\end{itemize} 
We compare these models to OpenAI's embeddings \citep{neelakantanTextCodeEmbeddings2022} through their API \footnote{accessed between October and December 2023}, using the model "text-embedding-ada-002" and to various non-finetuned models: pretrained-encoders Bert \cite{devlinBERTPretrainingDeep2019} and Roberta \cite{liuRoBERTaRobustlyOptimized2019}, pretrained decoders Mistral 7B-v0.1 \cite{jiangMistral7B2023}, LLaMA 1 \cite{touvronLLaMAOpenEfficient2023} and LLaMA 2 \cite{touvronLlamaOpenFoundation2023}, as well as the Pythia models \cite{bidermanPythiaSuiteAnalyzing2023}.
For these models, the embeddings are obtained by "mean pooling" except when specified otherwise \citep{reimersSentenceBERTSentenceEmbeddings2019}, i.e they are obtained by averaging the embeddings of each token at the last layer of the model. To reduce the dimension of text embeddings, we use a PCA with 30 components if not specified otherwise. 
We study this choice in \autoref{simple_pipeline} and show
that it is indeed a good default.
Finally, we compare these models to the simpler word model Fasttext \cite{bojanowskiEnrichingWordVectors2017}.

Table \ref{table:models} lists all the specific models that we investigate, with their major characteristics.

\paragraph{Substring based approaches}
For comparison, we also use character-level approach based on substrings.
We use scikit-learn's TfidfVectorizer \footnote{which is equivalent to the perhaps better known CountVectorizer, followed by a TfidfTransformer} to create an embedding based on
the occurrence of character-level ngrams. This embedding, which has the drawback of being
very high-dimensional, is then handled like embeddings from language
models. A more advanced and faster model we use is the MinHashEncoder
\citep{cerdaEncodingHighcardinalityString2022}, available through the
package Skrub \cite{skrub2023}, which takes advantage of the min-hash
approximation of the Jaccard to build encodings whose $L0$ distances are
approximations of the Jaccard of their ngrams sets. By default, we use 30
components to reach the same dimension that the reduced language model
embeddings.

\bigskip
\section{Results: gauging embeddings from simple to complex}
\label{results}

\subsection{Sophisticated string embeddings matter}

\begin{figure}
    \centering
    \includegraphics[clip, trim={1.5cm 0 1.5cm 0}, width=\linewidth]{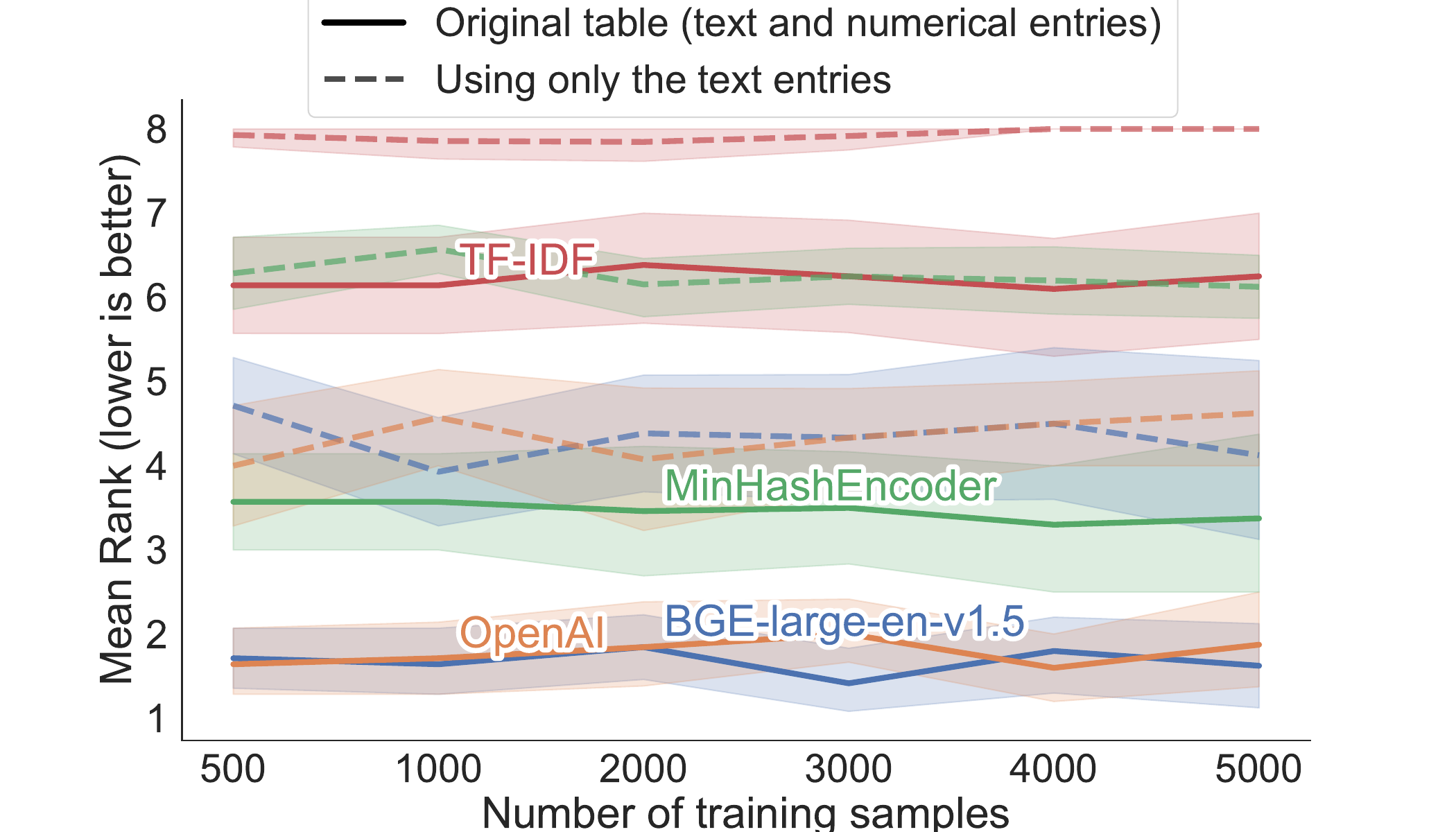}
    \caption{\textbf{Analytics: more sophisticated embedding improve performance} across varying training sizes using sklearn's GradientBoostingClassifier. The ranks are computed across both settings (predicting from text + numerical entries and predicting only from text entires), but not across sample size, and averaged on 14 datasets.}
    \label{mean_rank}
\end{figure}

We benchmark the performance of simple pipelines using entry embeddings
in our two settings: prediction and many-to-one fuzzy join.
\paragraph{Prediction}
Figure \ref{mean_rank} shows the performance of two language model
embedding methods (OpenAI's ada-002 and BAAI's BGE-large-en-v1.5)
compared to using Skrub's MinHashEncoder and sklearn's
TfidfVectorizer \footnote{for the TfidfVectorizer we only display the best set of parameters we found, which is using a ngram range of (2, 3) on characters. We varied the ngram range among \{(1, 2), (1, 3), (2, 3), (2, 4)\} both on characters and words, and with and without TFidf transformation.},
two string-based models described in \ref{text_embeddings}. On average
across our 14 analytic tasks and across all
training sizes from 500 to 5000, more sophisticated embeddings improve
task performance: the MinHashEncoder outperforms TF-IDF vectorization,
and OpenAI's text embedding is best. This order is preserved whether we
consider only the text columns, or all columns for the analysis. Jointly
modeling text and numerical columns brings a notable benefit, which
underlines the benefit of representing text with vectors of numbers.

\begin{figure}
    \centering
    \includegraphics[clip, trim={.2cm 0 .4cm 0}, width=\linewidth]{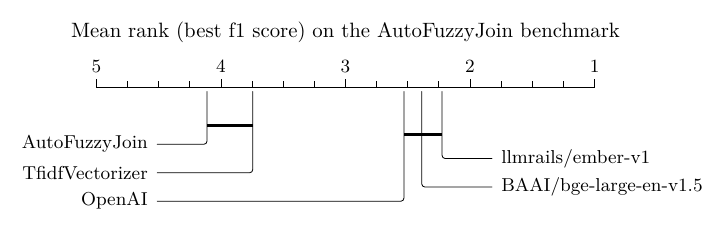}
    \caption{\textbf{Fuzzy join: pretrained language models improve upon
string distances} Average rank on AutoFuzzyJoin many-to-one unsupervised
join benchmark using different encoders. For each method expect
AutoFuzzyJoin, we use the 1-Nearest-Neighbor from Scikit-learn, adapting
the fuzzy join implementation in Skrub. The best F1 score is taken for
each dataset and method from 7 candidates by varying the precision /
recall tradeoff parameter (threshold for Nearest-Neighbor, target for
AutoFuzzyJoin) between 0.3 and 0.9%\Edouard{Est-ce que tu connais un autre
%papier qui utilise ctte représentation? Quand je vois les ordes de
%grandeurs des delta, j'ai envie de voir une notion de variance pour
%comprendre la significance de cette figure}%
%\textbf{Gael: c'est assez classic (cela s'appelle un "critical difference
%diagram" et c'est recommendé comme une "good practice"), un exemple de
%papier: \url{https://arxiv.org/pdf/2305.02997.pdf}}
}
    \label{fig:join}
\end{figure}

Table \ref{table:models} shows the performance of all the models we
evaluate, and in particular the mean difference with Skrub's
MinHashEncoder. We can see that, on average across the 14 tasks, all the
language models that we investigate improve upon the MinHashEncoder. 

\paragraph{Fuzzy Join}

% In the realm of tabular data processing, prediction is just one piece of the puzzle. In fact, data cleaning often consumes the majority of an analyst's time. This section explores the potential of language model embeddings to assist with this crucial step, focusing of the task of joining multiple tables. 
Figure \ref{fig:join} shows the performance of a simple approach: utilizing language model embeddings as input for a 1-Nearest-Neighbor algorithm. Using this simple pipeline with three strong language embedding models (see \ref{text_embeddings}), we show that this baseline outperforms the AutoFuzzyJoin algorithm \cite{liAutoFuzzyJoinAutoProgramFuzzy2021}, as well as a 1-Nearest-Neighbor using sklearn's TfidfVectorizer, on the 50 datasets benchmark from \cite{liAutoFuzzyJoinAutoProgramFuzzy2021} (see \ref{fuzzy_join}).

\bigskip
\subsection{Two different regimes: dirty categories and diverse entries}

\begin{figure}[!tb]
    {\small\sffamily ROC-AUC gain from OpenAI embeddings over MinHashEncoder}\\
    \includegraphics[clip, trim={0 0 0 2.1cm}, width=\linewidth]{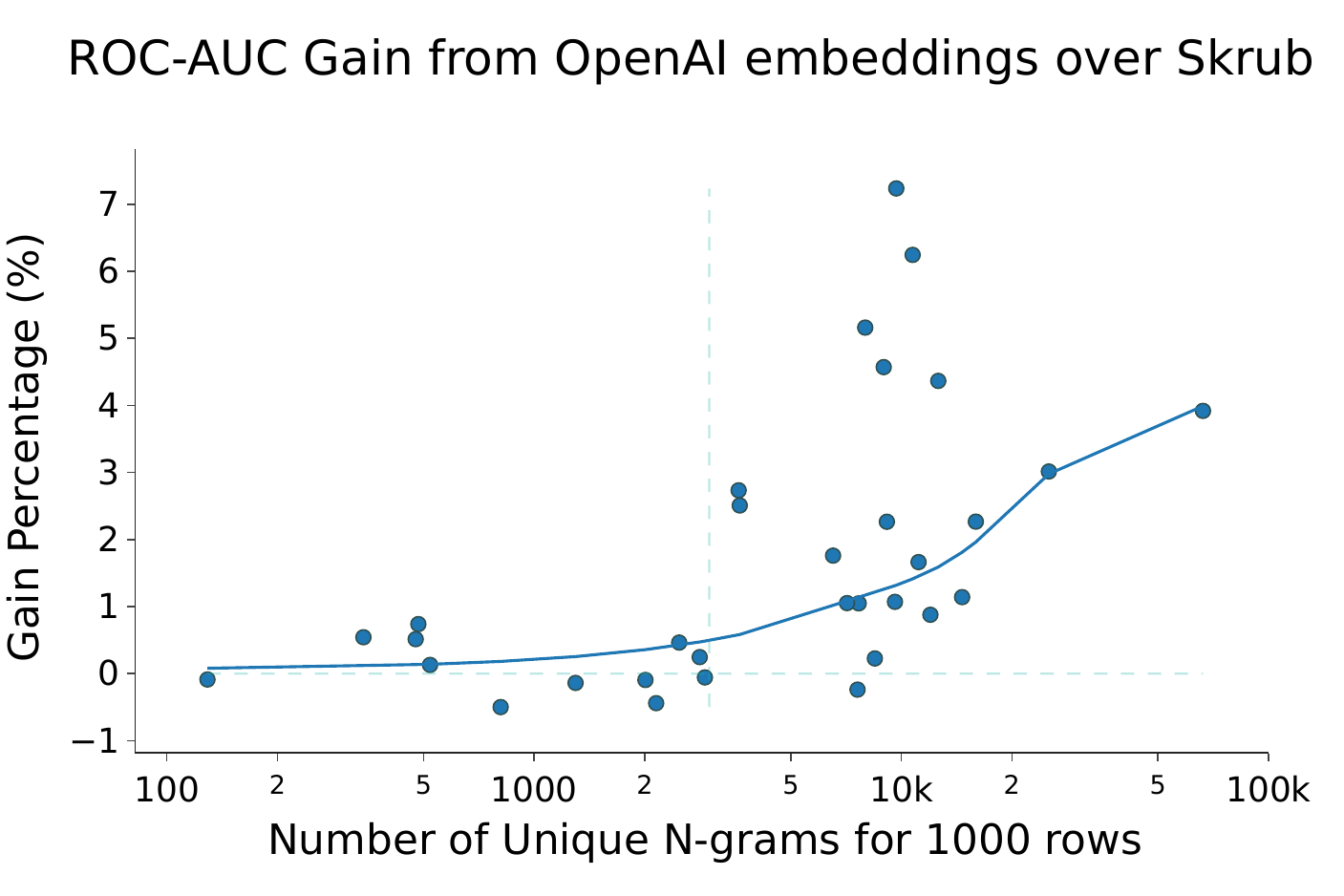}%
    \llap{\raisebox{.5\linewidth}{\parbox{.38\linewidth}{\sffamily\small
	as a function of diversity of strings
    }}\hspace*{.49\linewidth}}
    \includegraphics[clip, trim={0 0 0 2.1cm}, width=\linewidth]{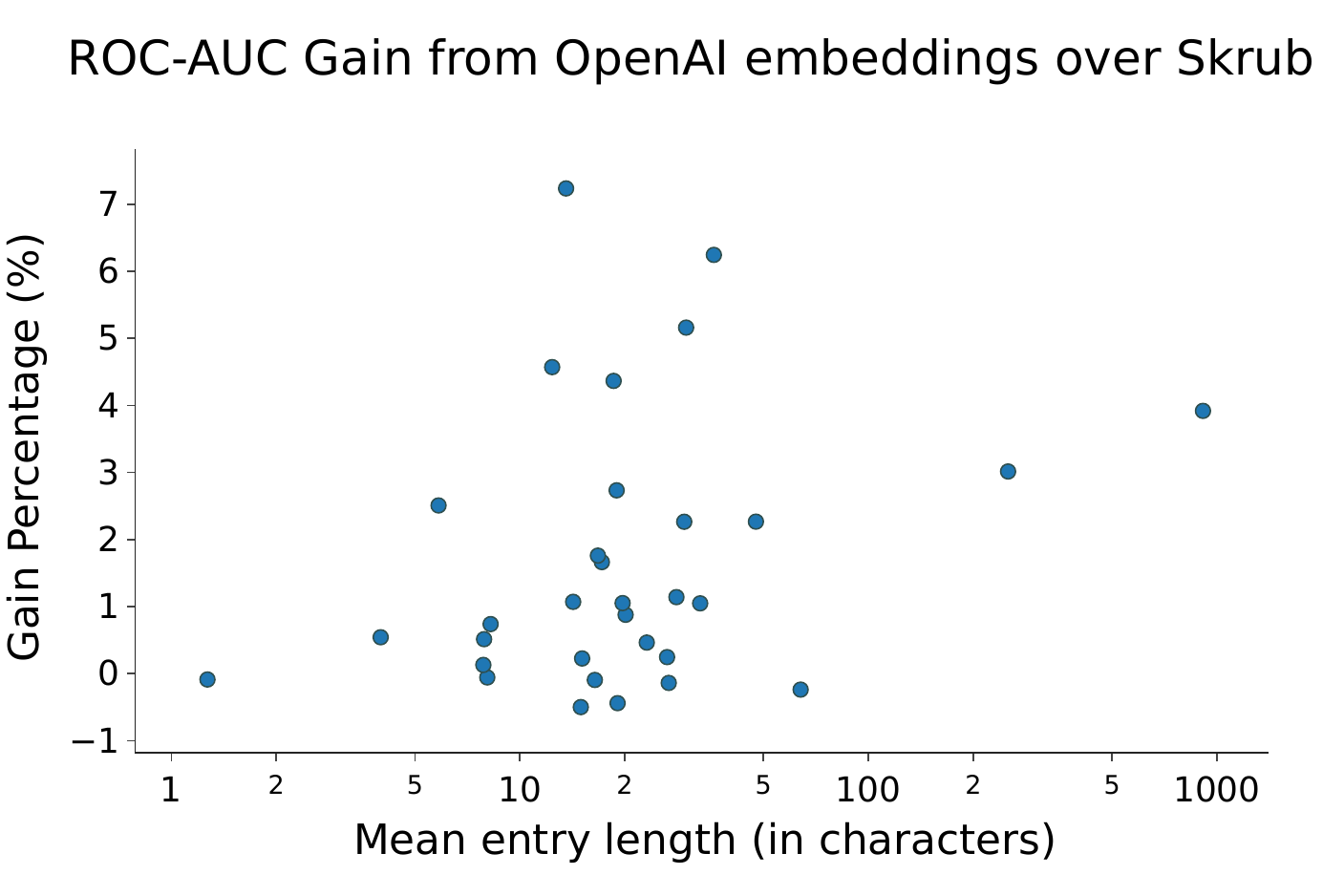}%
    \llap{\raisebox{.48\linewidth}{\parbox{.26\linewidth}{\sffamily\small
	as a function of string length
    }}\hspace*{.62\linewidth}}
        \caption{\textbf{The number of unique ngrams per row predicts the
gain better than the length of the text entries.} 
For every useful text column of 14 datasets, we compute the gain from replacing Skrub's MinHashEncoder encoding of this column with OpenAI embeddings (while keeping the other columns encoded as before, see setup). The experiment is done at a train size of 1000. 
Each point corresponds to a text entry from one of our datasets. The ngrams are based on characters, and computed between lengths 2 and 4.
}
        \label{roc_auc_gain}
        \label{fig:mean_length}
    \label{fig:which_column}
\end{figure}

\begin{figure}[tb!]
    \centering
    \includegraphics[width=\linewidth]{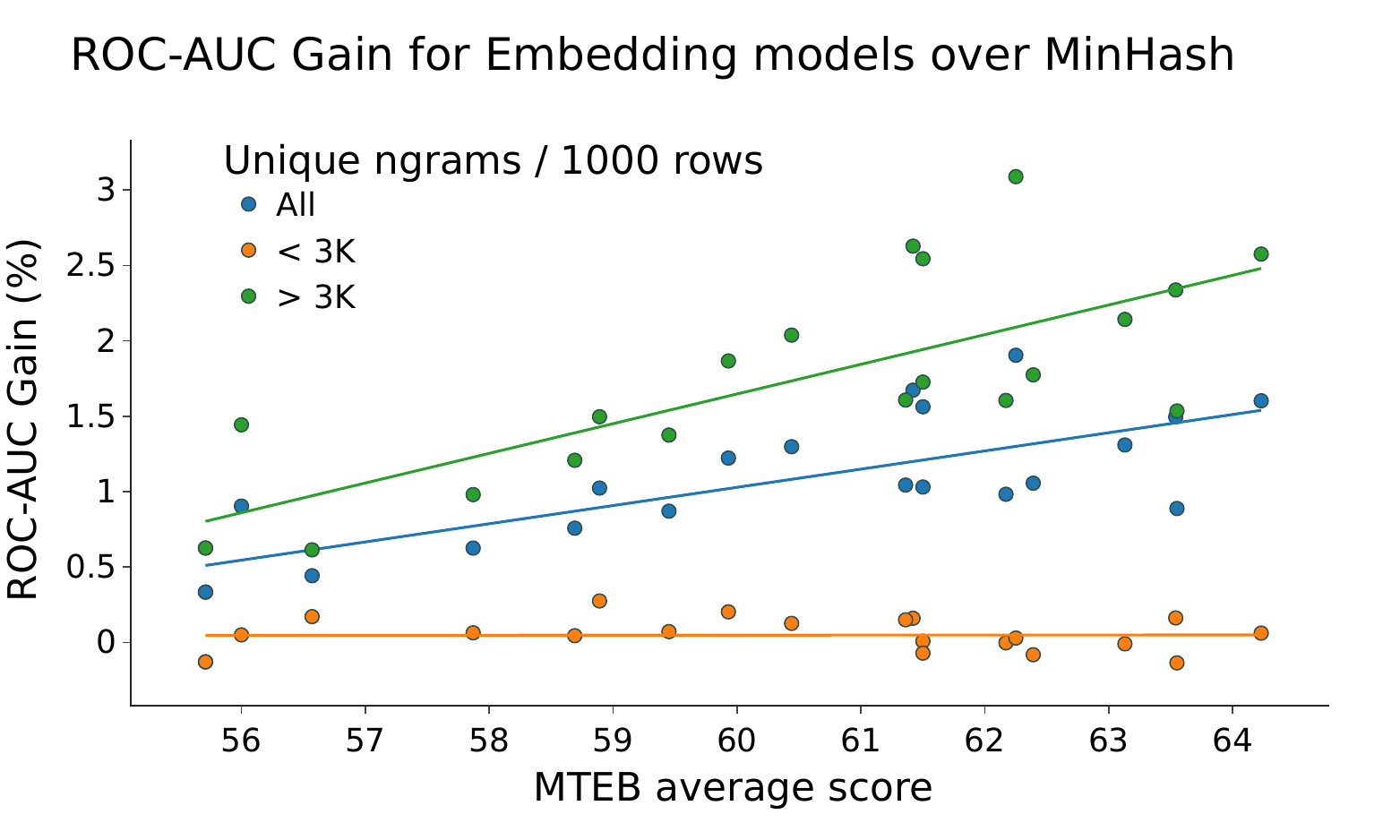}
    \caption{\textbf{Being better on classical embedding tasks translates to being better on tabular analytics}. But this is only the case in the \textit{diverse entries} regime, where the number of unique ngrams in the column is large enough. The gain is computed for each column by replacing the MinHash encoding by a language model embedding (+ PCA), and averaged accross columns (12 for ngrams < 3K, 20 for ngrams > 3K).}
    \label{fig:mteb_vs_roc_auc}
\end{figure}

\begin{figure*}[t!]
\centering
  \includegraphics[width=\linewidth]{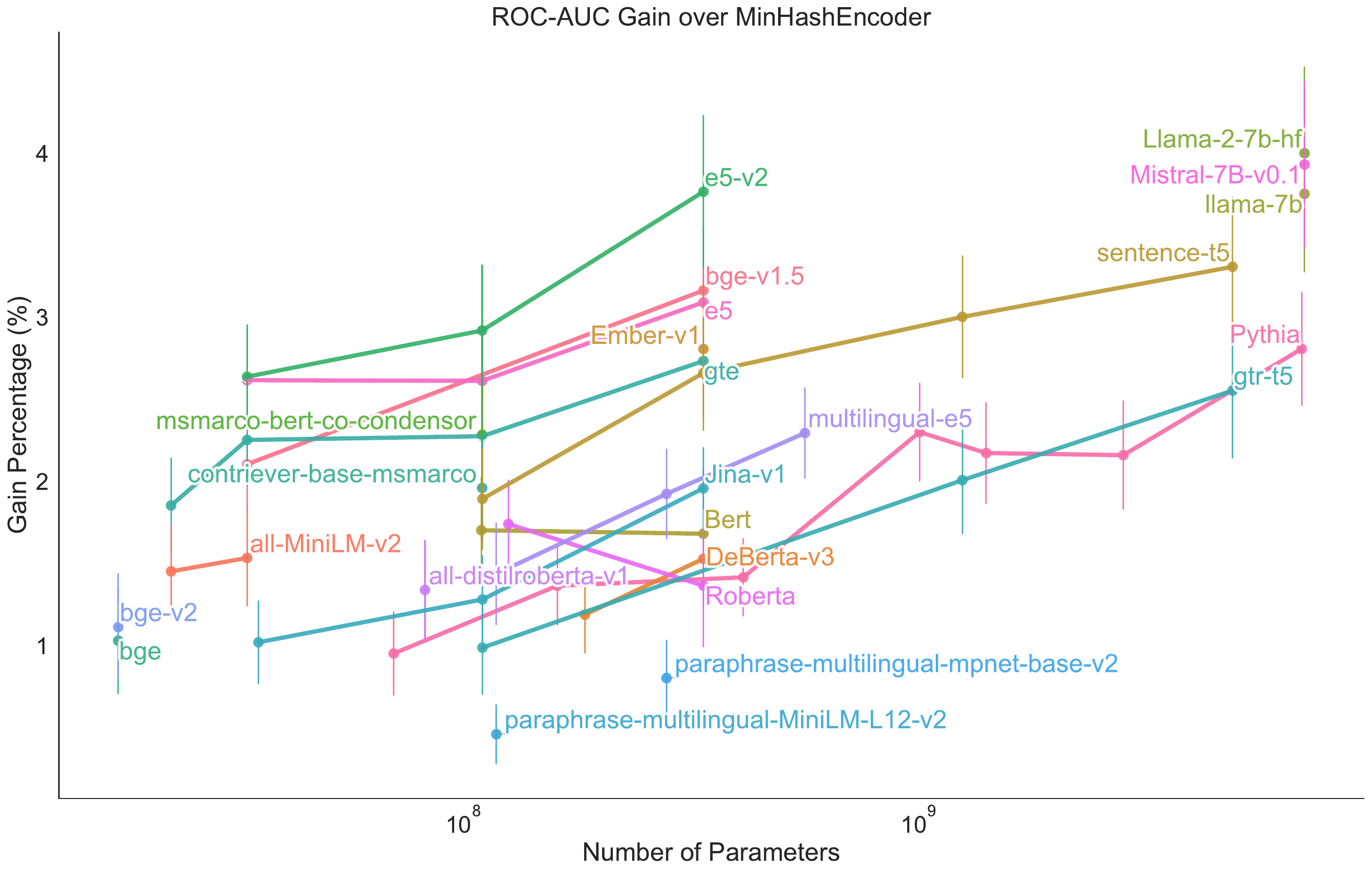}
  \caption{\textbf{Bigger is better, with diminishing returns} -- We vary the
model used to encode text features in place of Skrub's MinHashEncoder.
The gain percentage is averaged across 14 datasets, and computed on a
train size of 1000, using a sklearn's GradientBoostingClassfier. The error bars represent the standard error (halved for readability). Both
text and numerical features are used. The models are taken from the top
of the MTEB benchmark, expect for the large decoder models (Pythia,
Mistral, Llama) which are included to represent recent LLMs, and Bert,
Roberta and DeBerta models. In a given model family bigger is better, but
across all, very large models (Llama, sentence-t5, Pythia,
mistral...) bring no significant benefits compared to e5-v2.}
    \label{scale}

\end{figure*}

\begin{figure}[tbh!]
    \centering%
    \includegraphics[width=\linewidth]{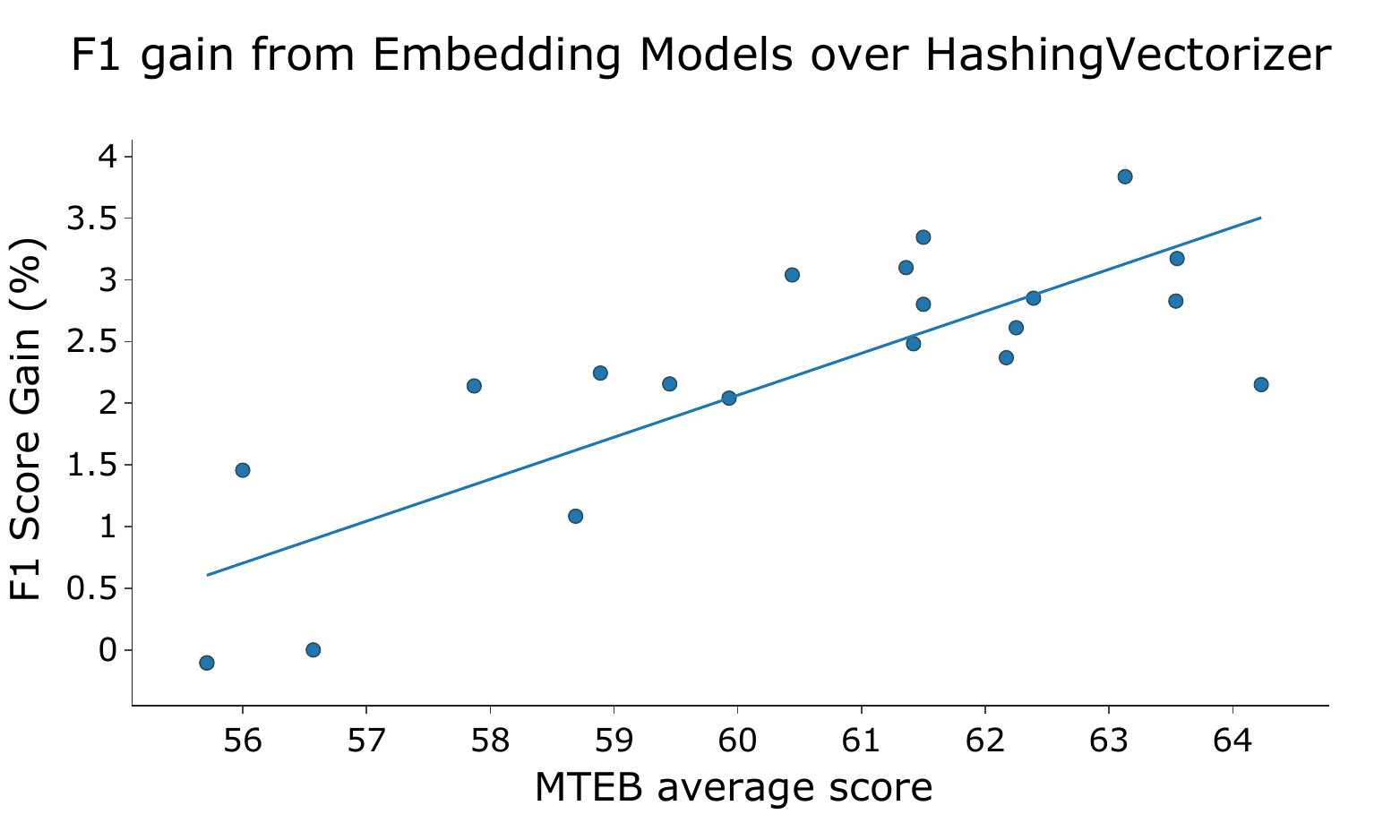}
    \caption{\textbf{Being better on classical embedding tasks translates
to being better for fuzzy joining.} The gain is averaged over 50
datasets, and computed by replacing the TfidfVectorizer by a
language model embeddings before the Nearest Neighbor. Note that all the
fuzzy join benchmark is in the "diverse entries" regime (more than 3000
unique ngrams for 1000 rows).\vspace*{-3ex}}
    \label{fig:f1_gain}
\end{figure}

Investigating the distribution of gains from using language model over
substring-based methods reveals that the benefits are unevenly
distributed.
In Figure \ref{roc_auc_gain}, we show the gain from using language model
encodings over MinHashEncoder on each useful column \footnote{i.e where
prediction is more than 0.5\% better when including this column with
either MinHashEncoder, OpenAI, or BAAI/bge-large-en-v1.5 embeddings over
dropping it} belonging to the datasets in our tabular analytics benchmark. We see approximately zero gain for slightly less than half of the
columns and significant gains for the other half. From the same Figure
\ref{roc_auc_gain}, we can separate the columns in two groups, based on a
simple metric, the number of unique ngrams in the column for 1000 rows
(computed on characters, between lengths of 2 and 4, for 1000 randomly sampled rows). This metric
captures how the diversity of strings grows as a function of number of
rows, revealing two regimes:
\begin{description}
    \item[dirty categories] columns where the number of unique ngrams is low, empirically below 3000 unique ngrams for 1000 rows. On these columns, it seems that using a language model brings little benefits over string-based approaches.
    \item[diverse entries] columns where the number of unique ngrams is high, empirically above 3000 unique ngrams for 1000 rows. On these columns, using language model embeddings brings significant improvement.
\end{description}

Table \ref{table:examples} shows examples of columns belonging to these
two categories. In contrast to our diversity metric, the length of the
text entries has little relationship with the gain from using language
model embeddings, as shown in Figure \ref{fig:mean_length}. Indeed a
column may contain strings that are both very short, but also very
diverse, as the {\tt artist name} column of the spotify dataset, for which using OpenAI's embedding over MinHashEncoder gives a ROC-AUC gain of 7.2\%.

\smallskip
\subsection{For diverse entries, using bigger, better models improves performance}

The above shows that for \textit{diverse entries}, using language model
embeddings improves over simpler string-based methods. This begs the
question, which embedding model should one use, among the enormous zoo of
available models? Benchmarks such as MTEB
\cite{muennighoffMTEBMassiveText2023} answer this questions for tasks
like passage retrieval or sentiment analysis. Do the
same tradeoffs apply to our case? Here text entries are much smaller than
typical texts, and the resulting embeddings of string entries are
combined with the other features of the tables before input to a
subsequent machine-learning model.

\begin{table*}[]
{\small
\rowcolors{2}{gray!25}{white}
\begin{tabular}{p{4cm}p{1.27cm}p{1.2cm}p{.7cm}p{1cm}p{.9cm}p{.9cm}p{1.7cm}p{2.02cm}}
\rowcolor{gray!50}
\textbf{Model} & \textbf{\!\!Parameters} & \textbf{Model type} & \textbf{Fine tuned} & \textbf{ROC-AUC Gain (\%)} & \textbf{Mean Rank (analytics)} & \textbf{F1 Gain (\%)} & \textbf{MTEB score (Average)} & \textbf{MTEB score (Classification)} \\
Llama-2-7b-hf                         & 7.0B       & Decoder    & No        & 4.0               & 12.64               & -30.0          & Unknown              & Unknown                     \\
Mistral-7B-v0.1                       & 7.0B       & Decoder    & No        & 3.93              & 13.0                & -20.6          & Unknown              & Unknown                     \\
e5-large-v2                           & 335.1M     & Encoder    & Yes       & 3.77              & 10.0                & 2.6            & 62.25                & 75.24                       \\
llama-7b                              & 7.0B       & Decoder    & No        & 3.75              & 14.0                & -29.4          & Unknown              & Unknown                     \\
sentence-t5-xxl                       & 4.9B       & Encoder    & Yes       & 3.31              & 17.29               & -1.7           & 59.51                & 73.42                       \\
bge-large-en-v1.5                     & 335.1M     & Encoder    & Yes       & 3.17              & 16.07               & 2.15           & 64.23                & 75.97                       \\
e5-large                              & 335.1M     & Encoder    & Yes       & 3.09              & 14.57               & 2.5            & 61.42                & 73.14                       \\
OpenAI Ada-002                        & Unknown    & Unknown    & Yes       & 2.86              & 19.79               & 2.8            & Unknown              & Unknown                     \\
pythia-6.9b                           & 6.9B       & Decoder    & No        & 2.81              & 21.21               & -24.8          & Unknown              & Unknown                     \\
ember-v1                              & 335.1M     & Encoder    & Yes       & 2.81              & 22.0                & 2.8            & 63.54                & 75.99                       \\
gte-large                             & 335.1M     & Encoder    & Yes       & 2.74              & 20.0                & 3.8            & 63.13                & 73.33                       \\
gtr-t5-xxl                            & 4.9B       & Encoder    & Yes       & 2.56              & 19.21               & 1.75           & 58.97                & 67.41                       \\
multilingual-e5-large                 & 559.9M     & Encoder    & Yes       & 2.3               & 26.14               & 2.8            & 61.5                 & 74.81                       \\
msmarco-bert-co-condensor             & 109.5M     & Encoder    & Yes       & 2.29              & 22.93               & 0.4            & 52.35                & 64.71                       \\
contriever-base-msmarco               & 109.5M     & Encoder    & Yes       & 1.96              & 29.36               & 1.5            & 56.0                 & 66.68                       \\
jina-embedding-l-en-v1                & 334.9M     & Encoder    & Yes       & 1.96              & 30.36               &                & Unknown              & Unknown                     \\
roberta-base                          & 125.0M     & Encoder    & No        & 1.74              & 31.5                & -31.9          & Unknown              & Unknown                     \\
bert-base-cased                       & 109.0M     & Encoder    & No        & 1.7               & 32.07               & -20.0          & Unknown              & Unknown                     \\
all-MiniLM-L12-v2                     & 33.4M      & Encoder    & Yes       & 1.54              & 30.29               & -1.2           & 56.53                & 63.21                       \\
deberta-v3-large                      & 335.0M     & Encoder    & No        & 1.53              & 36.86               &                & Unknown              & Unknown                     \\
Fasttext (cc-en)                      &            &            &           & 1.53              & 38.64               &                & Unknown              & Unknown                     \\
all-distilroberta-v1                  & 82.0M      & Encoder    & Yes       & 1.34              & 32.21               & -1.7           & Unknown              & Unknown                     \\
bge-micro-v2                          & 17.4M      & Encoder    & Yes       & 1.11              & 36.21               & 0.0            & 56.57                & 68.04                       \\
bge-micro                             & 17.4M      & Encoder    & Yes       & 1.03              & 30.79               & -0.1           & 55.71                & 66.35                       \\
paraphrase-multilingual-mpnet-base-v2 & 278.0M     & Encoder    & Yes       & 0.8               & 39.43               & -3.5           & Unknown              & Unknown                     \\
paraphrase-multilingual-MiniLM-L12-v2 & 117.7M     & Encoder    & Yes       & 0.46              & 42.64               & -9.3           & Unknown              & Unknown                     \\
Skrub MinHashEncoder                  &            &            &           & 0.0               & 44.57               &                & Unknown              & Unknown                              
\end{tabular}
}
\caption{\textbf{Performances of various models averaged across 14
datasets} for analytics (ROC-AUC gain and mean rank), and for 50 datasets for fuzzy-join (F1 gain). Performances are computed for a sample size of 1000, and using our default pipeline (PCA with 30 components, GradientBoostingClassifier) for analytics. If a model comes as a suite of models, we only show the best performing one.}
\label{table:models}
\end{table*}

\paragraph{Comparison to embeddings benchmarks}
Nonetheless, Figure \ref{fig:mteb_vs_roc_auc} shows that being better on the MTEB benchmark (on the average of the 56 tasks in the benchmark) quite directly translate to better performances on our tabular analytics tasks, in the \textit{diverse entries} regime. In the \textit{dirty categories} regime, in contrast, we see no gain from using better models.

The fuzzy join benchmark described in \ref{fuzzy_join} only contains columns in the \textit{diverse entries} regime, with more than 3000 unique ngrams for 1000 rows. Quite logically, we also observe in Figure \ref{fig:f1_gain} that being better on the MTEB benchmark translates to being better on this benchmark as well.

\paragraph{Bigger is better}
Embedding diverse entries of tables thus also follows the 
``bigger is better'' scaling behavior described across a range of natural
language tasks \cite{kaplanScalingLawsNeural2020}. For a given family of
models, figure \ref{scale} shows clear gains from increasing the model
size. Existing pre-trained model families enable us to investing this
trend for fine-tuned encoder models, such as e5
\cite{wangTextEmbeddingsWeaklySupervised2022}, but also decoder models with
Pythia \cite{bidermanPythiaSuiteAnalyzing2023} models. For a given family
we do not observe a plateau as we increase the model size.

In Table \ref{table:models}, we also see that among the biggest models we
evaluate, Mistral \cite{jiangMistral7B2023} and LLaMA 1
\cite{touvronLLaMAOpenEfficient2023} and 2
\cite{touvronLlamaOpenFoundation2023} are on top of our leaderboard,
despite being decoder models not finetuned for sentence similarity. This
suggests that our pipeline will be able to benefit from both current and
future advances in language models. This analysis could be extended to
other features known as being important for large language models, such
as the training and finetuning data quantity. The worse performance of Pythia 6.9B is perhaps due to being trained on 300B tokens compared to 1 and 2T for LLaMA 1 and 2.
%TODO: find citation saying that base models are not good

\paragraph{Finetuning}
While a given model family exhibit a ``bigger is better'' scaling
behavior on our tasks, finetuning the model for sentence embeddings is as
important, maybe more. Indeed, in \ref{table:models} and Figure \ref{scale}, we see that small finetuned models like bge or e5 arrive at close or better performances than the largest models in our table while being an order of magnitude smaller (330M vs 7B parameters). 
Moreover, we see in Figure \ref{scale} that a better and newer finetuning procedure translates to bigger gain on the tabular analytics task, as can be seen comparing the different versions of a finetuned model like e5.

\begin{figure}
  \centering
  \includegraphics[width=\linewidth]{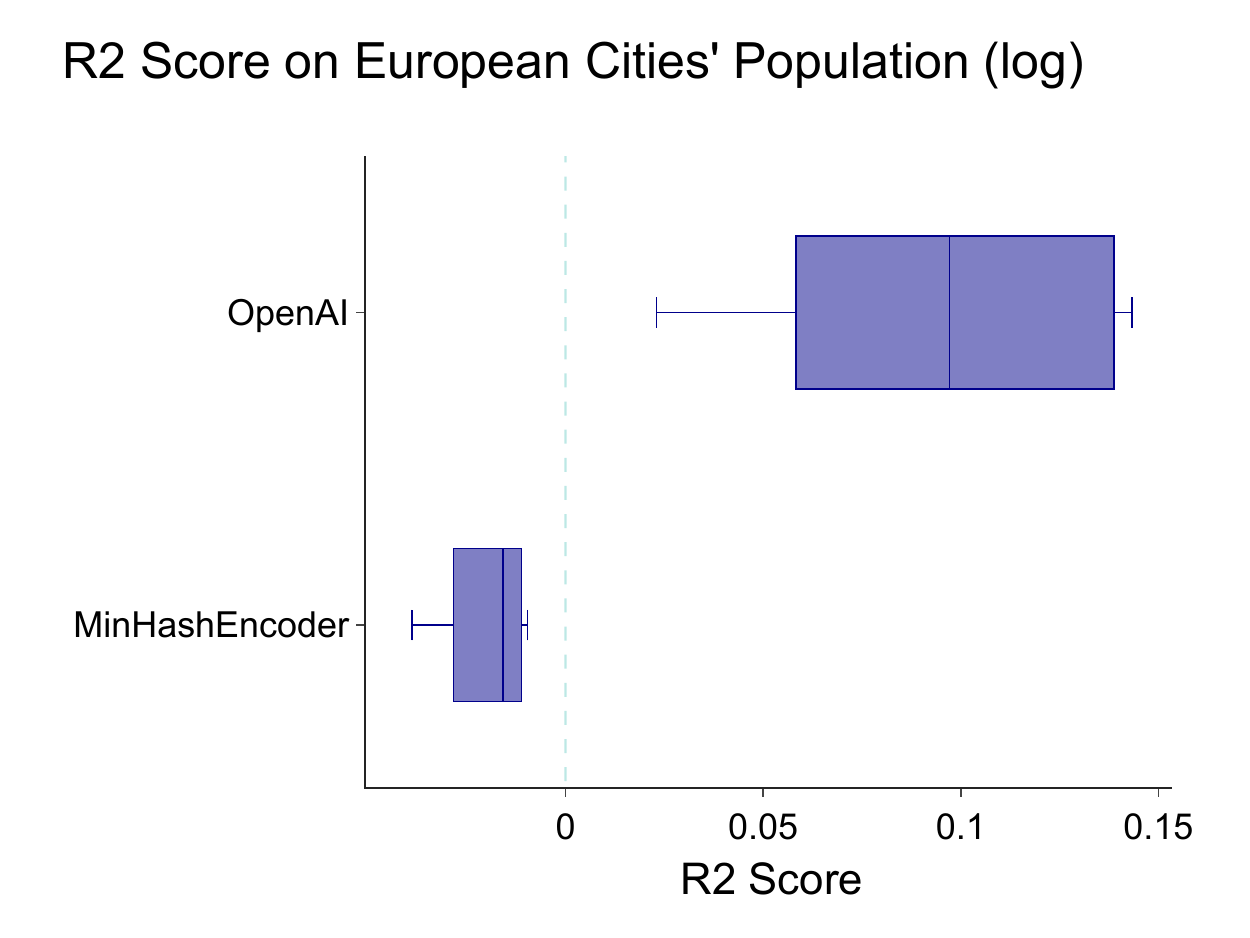}
  \caption{\textbf{Language models embed background knowledge} -- Comparison of the performance of Skrub's MinHashEncoder and OpenAI's embeddings for predicting the (log) population from the city and country names. This task is designed so that background knowledge is necessary for successful prediction: a city's population cannot be concluded from its name. As typical city size vary across countries, we design the train and test splits so that cities from a same country appear either in the train or the test set, but not in both.}
      \label{european_cities}
\end{figure}

\subsection{Language model can extract valuable knowledge from text features}
We hypothesize that the performance gains from using language models to encode text entries come from the background knowledge contained in these models \citep{gurneeLanguageModelsRepresent2023}. We provide some evidence for this claim in Figure \ref{european_cities}, where the task is to predict the population of Europeans cities (with more than 10K inhabitants) from their name, and the names of their countries. 
Here, to ensure that the learner does not simply recognize the country of
a city from its name --as city sizes differ between countries--
the split between the train and test set is done using sklearn's GroupKFold, such that the same country cannot appear both in the train and test set. We see that this makes it very hard for substring-based approach, as using Skrub's MinHashEncoder leads to performance akin to random chance. On the contrary, using the OpenAI embedding, we are able to retain decent performances, suggesting that we are actually using the population knowledge contained inside the embedding.

\subsection{A solid default pipeline}
\label{simple_pipeline}

\begin{figure}[t]
    \centering
    \includegraphics[width=\linewidth]{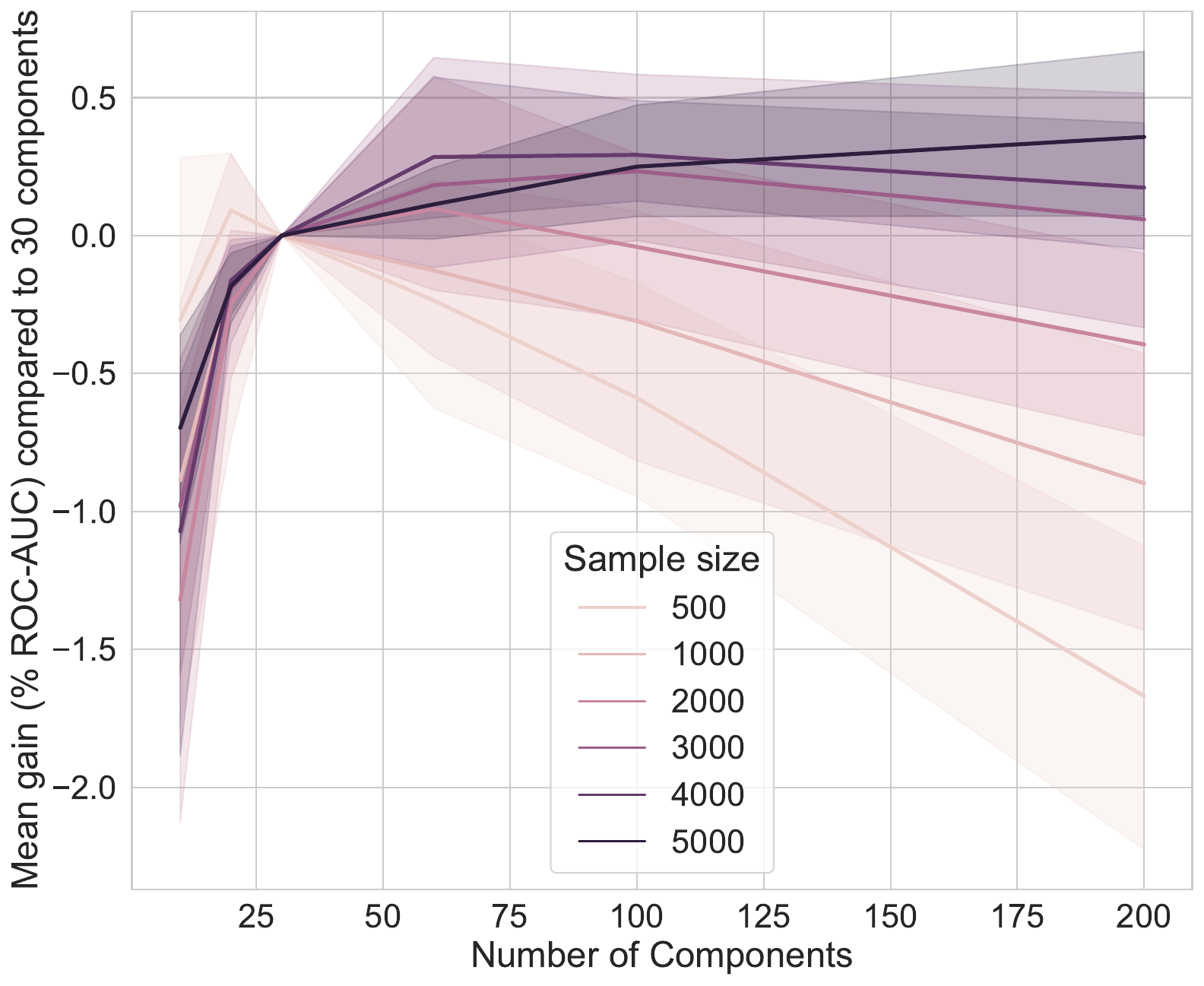}
    \caption{\textbf{Choosing 30 components in the PCA is a reasonable choice.} Comparing the performance of various choices of number of components for the PCA using OpenAI's embeddings and evaluated using a GradientBoostingClassifier. While bigger number of components seems to improve slightly over 30 for bigger train sizes, 30 seems optimal for below 2000 train sizes, and close to optimal until 5000.}
    %TODO sample size instead of train_size
    \label{fig:varying_dim}
\end{figure}

\begin{figure}[t]
    \centering
    \includegraphics[width=\linewidth]{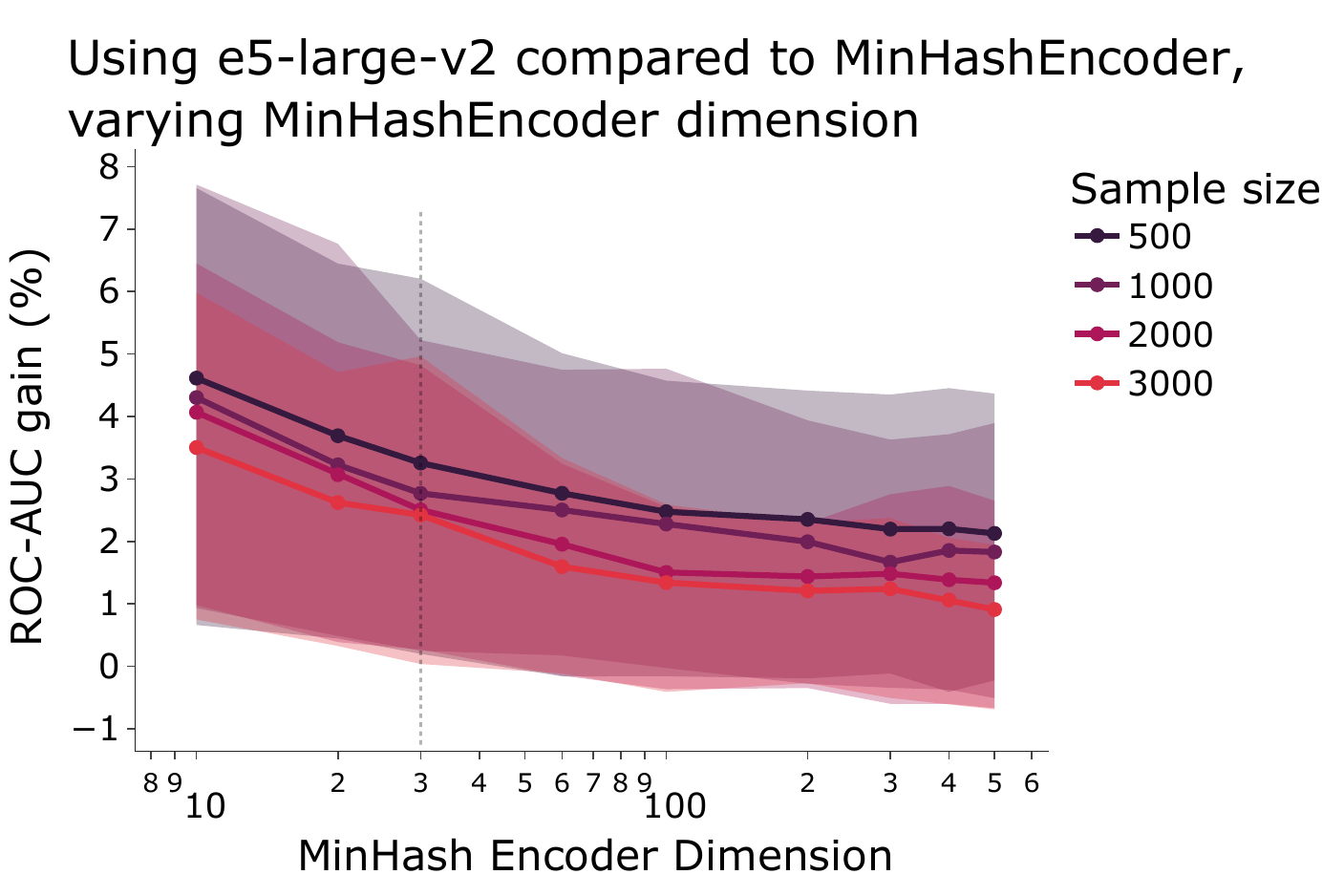}
    \caption{\textbf{Language models stay superior when increasing the MinHashEncoder dimension} while keeping the language model embeddings to 30 dimension, though the gap narrows. We note that increasing this dimension leads to higher downstream compute cost.}
    \label{fig:minhash_varying_dim}
\end{figure}

In this section, we check the robustness of our default pipeline using a series of ablations. The purpose is twofold: checking that the results of our experiments are not tainted by subpar settings, and guiding practitioners toward a simple yet effective pipeline. We recall that our default pipeline consists of encoding text entries with a language models, reducing the dimension of these embeddings with a Principal Component Analysis with 30 components, concatenating the results with the numerical features, and training a GradientBoostingClassifier on the result.

In Figure \ref{fig:varying_dim}, we vary the number of components of the Principal Component Analysis used to reduce the embeddings dimension, and display the mean gain compared to using a dimension of 30, our default. We see that a dimension of 30 seems optimal until a sample size of 2000, and very close to optimal for bigger sample size until 5000 (the biggest size in our experiments). 

In our paper, we kept the embedding dimension constant for all methods. In Figure 9, we vary the MinHashEncoder dimension while keeping the language model embedding dimension to 30 (using PCA). We see that up to 500, language models stay superior, with only 30 dimensions. We note that increasing the embedding dimension can leads to significantly higher downstream compute cost. Depending on how much embeddings can be reused, the higher cost of language model can be offset by using a smaller dimension.

\begin{figure}
    \centering
    \includegraphics[width=\linewidth]{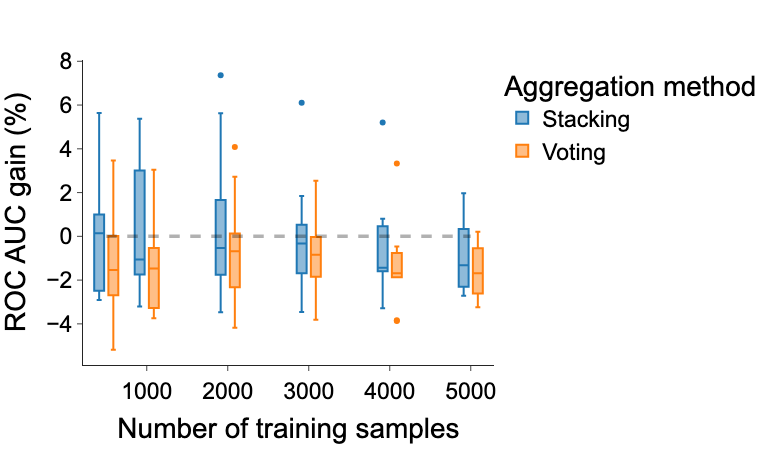}
    \caption{\textbf{Using a linear model on the embeddings underperforms our simple tree-based pipeline.} ROC-AUC gain of using an ensemble of GradientBoostingClassifier on numerical features and a LogisticRegression on (OpenAI) embeddings (without dimensionality reduction), followed by either a stacking (using LogisticRegression) or (soft) voting using sklearn's StackingClassifier or VotingClassifier, compared to our usual pipeline of using a PCA (dim 30) on embeddings, followed by a GradientBoostingClassifier on all features.}
    \label{fig:ensembling}
\end{figure}

Next, we study whether ensembling different models for text embeddings and numerical features beats our simple pipeline. Indeed, using a tree-based model on language model embeddings is unusual, and some work have shown that features are often linearly encoded in language models activations \cite{gurneeLanguageModelsRepresent2023}. To this aim, we ensemble the prediction of a GradientBoostingClassifier trained on numerical features and a LogisticRegression trained on the text embeddings (without dimensionality reduction), and compute the mean ROC-AUC gain (accross datasets) compared to our pipeline. The ensembling is done either using scikit-learn's VotingClassifier, i.e averaging the probability of each class, or using scikit-learn's StackingClassifier, i.e training a LogisticRegression on the output of both ensembled models. As we can see in Figure \ref{fig:ensembling}, both embedding methods fail to improve upon our baseline on average. We do note however that on certain datasets, these methods bring improvements.

\bigskip
\section*{Conclusion}
\label{sec:conlcusion}

%discuss biases and limitations somewhere

\paragraph*{Rules of thumb}
A thorough benchmark of embedding string entries for various data
processing applications highlights trends precious for data engineering.
These can be distilled in simple guidelines, good defaults to save
practitioners time. First, it is useful to distinguish two kind of string
columns: \emph{dirty categories} with a low diversity across strings (for
1\,000 rows, no more than 3\,000 unique character-level $n$-grams with $n
\in \{2, 3, 4\}$), and \emph{diverse entries}. For dirty categories,
lightweight string representations as the MinHashEncoder
\cite{cerdaEncodingHighcardinalityString2022,skrub2023} suffice. For
diverse entries, borrowing language models from recent NLP developments
brings much benefits. Here, bigger and more advanced language models to
represent text entries in tables capture better knowledge useful for
prediction and preprocessing on tables. For these columns, the findings from text embedding 
in natural language models carry over: larger models, fine-tuned to sentence-comparison tasks, bring benefits to analytic and entity resolution tasks. 
In particular, they markedly outperform word embeddings such as FastText which are currently often used as a default solution.
Larger models come with increased computational burdens, and it can be
useful to favor well fine-tuned models. To date, e5 (v2) \cite{wangTextEmbeddingsWeaklySupervised2022}
stands out as an excellent compromise.

\paragraph*{Future work}
Given a large database, better representations can be probably be
obtained by adapting models to the database. However, this will increase
markedly the computational and operational costs.
The simple pipeline that we studied
can easily be scaled to large datasets: the embedding complexity is
linear with the number of records, and embeddings can be computed only once. Furthermore, progress in language model inference \cite{timbersTransformerInferenceTricks2023} \cite{shengFlexGenHighThroughputGenerative2023} can make the embedding computation faster and cheaper. 
An interesting avenue of research would be to study whether the
particular background information we need for tabular analytics can be
accessed without running the whole language model, as it has been
observed that better information can be extracted from earlier layers in
large language models \cite{mengLocatingEditingFactual2023}.

%\cleardoublepage
\bigskip

\bibliographystyle{IEEEtran}
\bibliography{workshop.bib}

\end{document}